\documentclass[runningheads]{llncs}
\usepackage[T1]{fontenc}
\usepackage{graphicx}
\usepackage{booktabs}
\usepackage{array}
\usepackage{amsmath}
\usepackage{amsfonts}
\usepackage{amssymb}
\usepackage[colorlinks,
            linkcolor=blue,
            anchorcolor=blue,
            citecolor=blue,
            urlcolor=blue]{hyperref}

\usepackage{CJKutf8}
\usepackage[encapsulated]{CJK}
\AtBeginDocument{\begin{CJK}{UTF8}{gbsn}}
\AtEndDocument{\end{CJK}}

\begin{document}
    \title{Syntax-Aware Complex-Valued Neural Machine Translation}
%
    \author{Yang Liu \and Yuexian Hou}

    \authorrunning{Y. Liu et al.}
%
    \institute{College of Intelligence and Computing, Tianjin University, Tianjin, China
    \email{\{lauyon,yxhou\}@tju.edu.cn}}
    \maketitle              
    \begin{abstract}
        Syntax has been proven to be remarkably effective in neural machine translation (NMT).
        Previous models obtained syntax information from syntactic parsing tools and integrated it into NMT models to improve translation performance.
        In this work, we propose a method to incorporate syntax information into a complex-valued Encoder-Decoder architecture.
        The proposed model jointly learns word-level and syntax-level attention scores from the source side to the target side using an attention mechanism.
        Importantly, it is not dependent on specific network architectures and can be directly integrated into any existing sequence-to-sequence (Seq2Seq) framework.
        The experimental results demonstrate that the proposed method can bring significant improvements in BLEU scores on two datasets.
        In particular, the proposed method achieves a greater improvement in BLEU scores in translation tasks involving language pairs with significant syntactic differences.

    \end{abstract}

    \section{Introduction}

    In recent years, neural machine translation (NMT) has benefited from the sequence-to-sequence (Seq2Seq) framework and attention mechanism \cite{bahdanau2014neural,luong-manning-2015-stanford,luong-etal-2015-effective,vaswani2017attention}.
    NMT has shown effective improvements over statistical machine translation (SMT) models in various language pairs.
    The attention-based encoder-decoder architecture encodes the source language sentence into a sequence of hidden real-valued vectors and then computes their weighted sum based on the attention mechanism to generate the target language sentence predictions in the decoder \cite{cho-etal-2014-properties}.

    \begin{figure}
        \includegraphics[width=\textwidth]{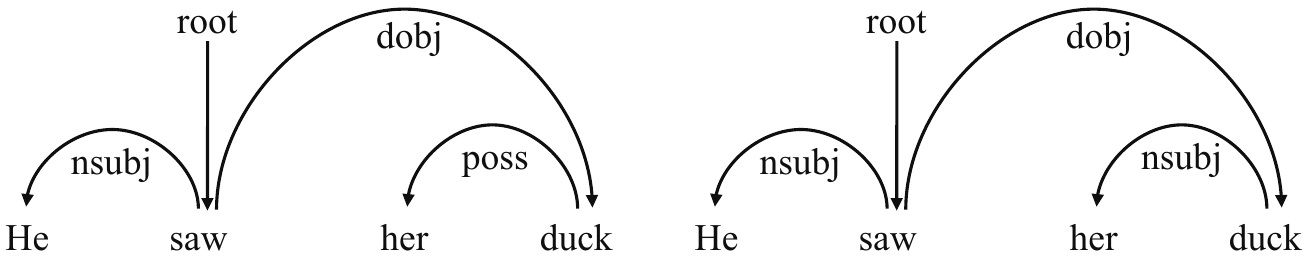}
        \caption{An example that illustrates how different syntactic dependency structures can lead to a change in the meaning of a sentence.} \label{fig:syntax_demo}
    \end{figure}

    Syntax information has been widely used and proven effective in SMT \cite{liu-etal-2006-tree,chiang-2007-hierarchical,williams-koehn-2014-syntax}, and researchers have also attempted to incorporate syntax information into NMT \cite{shi-etal-2016-string,ijcai2017p584,hashimoto-tsuruoka-2017-neural}.
    Several studies have demonstrated the significance of syntax information, which is also briefly discussed in this work.
    For example, in the sentence ``He saw her duck'',
    according to the syntactic analysis results shown in Figure \ref{fig:syntax_demo},
    when ``duck'' is used as an attributive of ``her'',
    it means ``a duck''.
    In this case, the dependency relationship between ``duck'' and ``her'' is \textit{poss} (possessive).
    However, when ``duck'' is used as the predicate of ``her'',
    it means ``to dodge'', ``to avoid'', or ``to duck down''.
    In this case, the dependency relationship between ``duck'' and ``her'' is \textit{nsubj} (noun subject).
    This suggests that syntactic information greatly affects the meaning of words.

\begin{figure}
        \includegraphics[width=\textwidth]{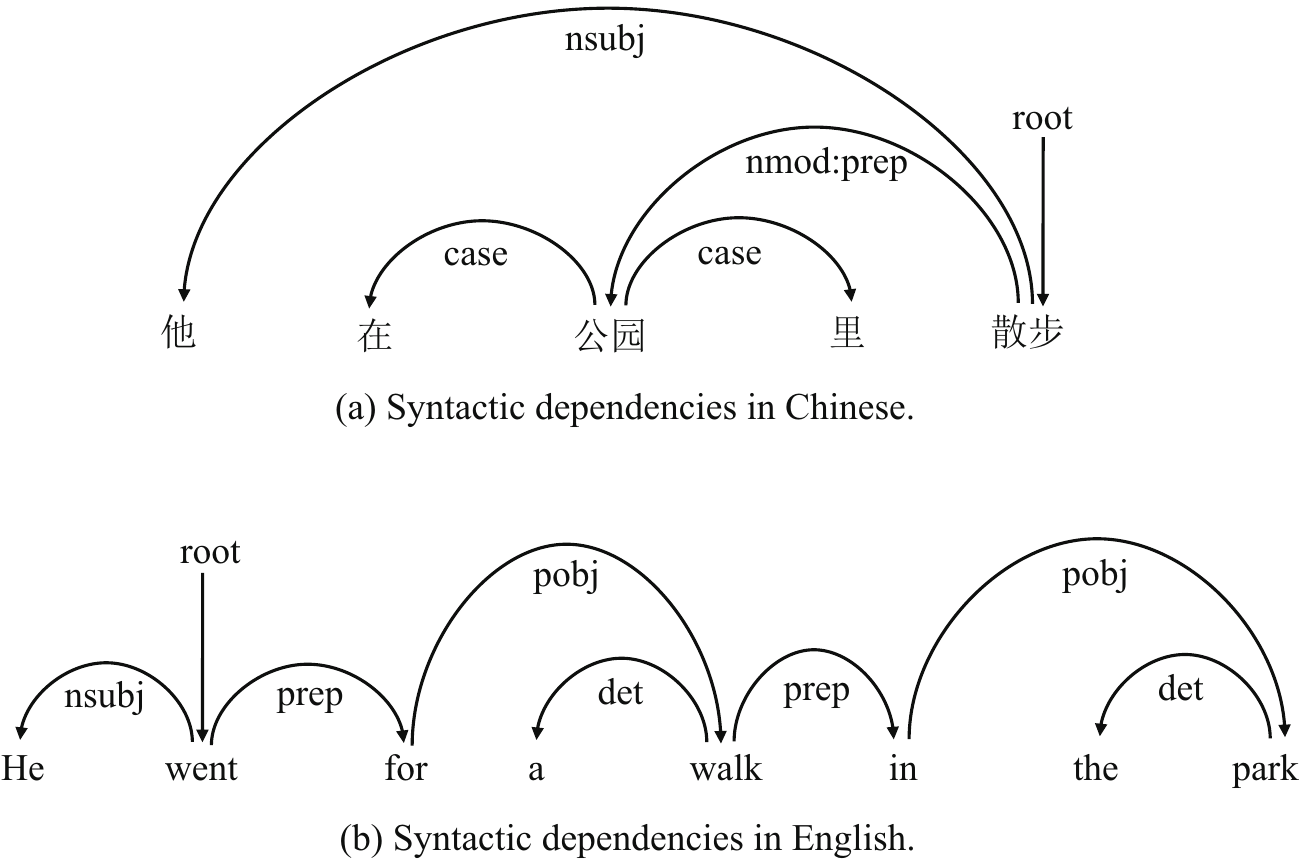}
        \caption{An example that illustrates how the same sentence meaning can have different syntactic dependency structures in different languages.}
        \label{fig:syntax_demo_differ_laug}
    \end{figure}

    A considerable number of researchers have integrated syntactic information into NMT to enhance translation performance.
    Most of these studies integrated syntax on the source side \cite{shi-etal-2016-string,hashimoto-tsuruoka-2017-neural}.
    However, in reality, the syntactic dependencies differ across different languages.
    Therefore, previous research lacked matching of syntactic dependencies across different languages.
    As shown in Figure \ref{fig:syntax_demo_differ_laug}, consider the Chinese sentence ``他在公园里散步。'' and its English translation ``He went for a walk in the park.''
    These two sentences convey the same meaning, which is describing someone walking in the park.
    However, from the perspective of syntactic analysis, they are not exactly the same.

    Recent work on recurrent neural networks and analysis of fundamental theory suggests that complex numbers may have a richer representational capacity \cite{trabelsi2018deep}.
    Many researchers have been enthusiastic about extending complex-valued neural networks (CVNNs), in which complex numbers are treated as either fixed parameters or meaningless additional parameters \cite{wang2019encoding,wang2019semantic}.
    However, these practices significantly constrain the generalization ability and interpretability of CVNNs.

    In this work, syntactic information from different languages is embedded into CVNNs.
    The complex-valued attention mechanism is used to score complex-valued hidden vectors, aiming to jointly learn the attention scores of both words and syntax.
    This approach does not depend on specific model architectures and can be applied to any existing Seq2Seq architecture.
    Our core idea is that the same word with different syntactic dependencies can have different meanings.
    This complex word embedding approach can alleviate to some extent translation errors caused by polysemy.
    We conducted experiments on Chinese-to-English and English-to-German translation tasks.
    Experimental results show that our approach is very effective, especially for long sentences.

    \section{Background}

    \subsection{Neural Machine Translation}
    A typical neural machine translation system is an attention-based Encoder-Decoder architecture, which models the conditional probability $p(y|x)$ by computing the source sentence $x_1,...,x_n$ to the target sentence $y_1,...,y_m$.
    An encoder is used to obtain a representation of the source sentence, and a decoder generates one target word at a time.
    The conditional probability can be decomposed as:
    \begin{equation}
        \log p(y|x) = \sum_{j=1}^m \log p(y_j|y_{<j},s)
    \end{equation}
    A recurrent neural network (RNN) is typically chosen as the encoder and decoder, including the long short-term memory (LSTM) or the gated recurrent unit(GRU).
    Taking RNN as an example, the hidden state $h_t$ at time $t$ is:

    \begin{equation}
        h_t = \mathrm{tanh}(W_{ih}x_t+b_{ih}+W_{hh}h_{t-1}+b_{hh})
    \end{equation}

    where $x_t$ is the input at time $t$, and $h_{t-1}$ is the hidden state at time $t-1$.
    The context vector $c_i$, which is used as input for the decoder, is computed as a weighted sum of these $h_t$:
    \begin{equation}
        c_i = \sum_{j=1}^{n} \alpha_{ij}h_j
    \end{equation}
    The weight $\alpha_{ij}$ for each hidden state $h_j$ is computed as:
    \begin{equation}
        \alpha_{ij} = \frac{\exp(e_{ij})}{\sum_{k=1}^{n}\exp(e_{ik})}
    \end{equation}
    where $e_{ij}=\mathrm{a}(s_{i-1}, h_j)$ is an alignment model, and $s_{i-1}$ represents the hidden state of the decoder at time $i-1$.

    \subsection{Complex-valued Neural Network}\label{complex_valued_neural_network}
    A complex number can be represented as $z=a+\mathrm{i}b$, where $a$ is the real part and $b$ is the imaginary part, and $\mathrm{i}$ is the imaginary unit.
    In CVNNs, a complex-valued vector can be represented as:
    \begin{equation}
        h=x+\mathrm{i}y
    \end{equation}
    where $x$ and $y$ are real-valued vectors.
    A complex-valued matrix can be represented as:
    \begin{equation}
        W=A+\mathrm{i} B
    \end{equation}
    where $A$ and $B$ are real-valued matrices.
    The calculation rule for the product of a complex-valued matrix and a complex-valued vector is as follows:
    \begin{equation}
        Wh=(Ax-By)+\mathrm{i}(Bx+Ay)
    \end{equation}
    It can be expressed in matrix notation as follows:
    \begin{equation}
        \begin{bmatrix}
            \Re (Wh) \\ \Im (Wh)
        \end{bmatrix}
        =
        \begin{bmatrix}
            A & \  -B \\ B & \  A
        \end{bmatrix}
        \begin{bmatrix}
            x \\ y
        \end{bmatrix}
    \end{equation}
    A complex-valued activation function (e.g. ReLU) can be represented as:
    \begin{equation}
        \mathbb{C} \mathrm{ReLU}(z) = \mathrm{ReLU}(\Re (z)) + \mathrm{i}\mathrm{ReLU}(\Im (z))
    \end{equation}


    \section{Proposed Approach}\label{sec:proposed_approach}
    \begin{figure}
        \includegraphics[width=\textwidth]{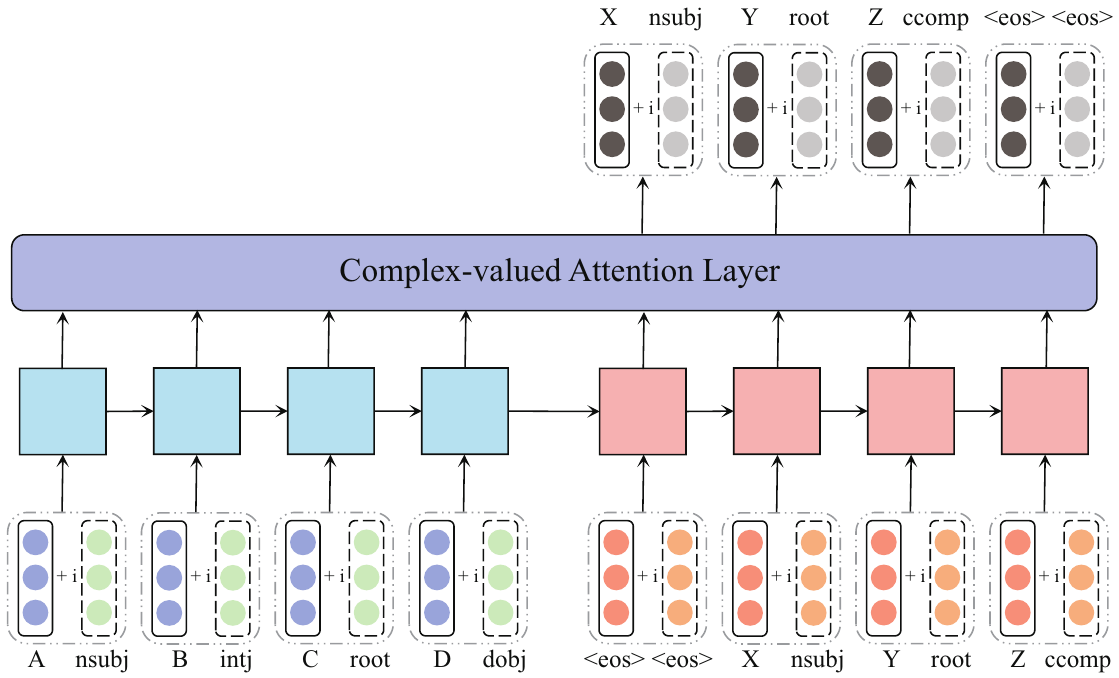}
        \caption{ Syntax-Aware Complex-Valued Neural Machine Translation.
        The words and their syntactic information are represented as complex-valued vectors with real and imaginary components respectively, and the entire network is computed using complex-valued operations.}
        \label{fig:synconmt}
    \end{figure}

    In this section, we will provide details about the approach we have proposed (the basic structure is shown in Figure \ref{fig:synconmt}).
    We refer to it as Syntax-Aware Complex-Valued Neural Machine Translation (SynCoNMT).
    For simplicity, we first agree to use the following notation to represent the general complex-valued form of a tensor:
    \begin{equation}
        \mathbb{C}(z) = \Re (z) + \mathrm{i} \Im (z)
    \end{equation}
    where $z$ can be a tensor of arbitrary shape.
    $\Re (z)$ and $\Im (z)$ respectively represent the real and imaginary parts of the complex tensor $z$.

    \subsection{Syntax Embedding}\label{syntax_embedding}
    In this work, we use CVNNs to represent words and syntactic information as the real and imaginary parts of complex-valued vectors, respectively.
    We refer to this embedding as Syntax Embedding (SE).
    The word vectors and dependency vectors are obtained from the word and dependency lookup tables, respectively.
    The SE of a word $w$, denoted as $\mathbb{C}(s_{ij})$, represents the meaning of the word $w_i$ in a specific syntactic dependency relation $d_j$, and is defined as follows:
    \begin{equation}
        \mathbb{C}(s_{ij}) =  w_i + \mathrm{i}d_j =  \Re (s_{ij}) + \mathrm{i}\Im (s_{ij})
    \end{equation}
    Furthermore, the meaning of word $w_i$ under the dependency relations $d_j$ and $d_k$ can be represented by two embeddings, $\mathbb{C}(s_{ij})$ and $\mathbb{C}(s_{ik})$, which have the same real part but different imaginary parts.

    \subsection{Complex-valued Neural Machine Translation}
    The SynCoNMT that we propose receives complex-valued syntactic embeddings (as described in section \ref{syntax_embedding}) and predicts the conditional probability distribution of the target word.
    The encoder and decoder use a complex-valued RNN architecture for computation.
    Specifically, the hidden state $\mathbb{C}(h_t)$ at time $t$ is:
    \begin{equation}
        \mathbb{C}(h_t) = \mathbb{C} \mathrm{tanh}(\mathbb{C}(W_{ih}x_t)+\mathbb{C}(b_{ih})+\mathbb{C}(W_{hh}h_{t-1})+\mathbb{C}(b_{hh}))
    \end{equation}
    The complex-valued context vector $\mathbb{C}(c_i)$ is computed as the weighted sum of the hidden states $\mathbb{C}(h_t)$ using complex-valued attention scores, as follows:
    \begin{equation}
        \mathbb{C}(c_i) = \sum_{j=1}^{n} \mathbb{C}(\alpha_{ij})\mathbb{C}(h_j)
    \end{equation}
    The weight $\mathbb{C}(\alpha_{ij})$ for each complex-valued hidden state $\mathbb{C}(h_j)$ is computed as:
    \begin{equation}
        \mathbb{C}(\alpha_{ij}) = \mathbb{C}\mathrm{Softmax}(\mathbb{C}(e_{ij}))
    \end{equation}
    where $\mathbb{C}(e_{ij})=\mathbb{C}\mathrm{align}(\mathbb{C}(s_{i-1}), \mathbb{C}(h_j))$ is a complex-valued alignment model that follows the complex-valued computation rules described in section \ref{complex_valued_neural_network}.
    The complex-valued attention scores weight both the word vectors and syntax vectors, allowing the decoder to focus on both semantic and syntactic information simultaneously.

    \subsection{Syntax-based Loss Function}
    The model we proposed simultaneously predicts words and syntax and uses the predicted words as the final translation result.
    We use two complex-valued fully connected layers to map the decoder's hidden state $s_i$ to the word and dependency space respectively:
    \begin{equation}
        \hat{y}_w = |\mathbb{C}\mathrm{Linear_w}(\mathbb{C}(s_{i}))|
    \end{equation}
    \begin{equation}
        \hat{y}_d = |\mathbb{C}\mathrm{Linear_d}(\mathbb{C}(s_{i}))|
    \end{equation}
    where $|\cdot|$ represents the computation of the modulus of each element in a complex-valued vector.
    We jointly compute the loss of predicting words and syntax dependencies using the cross-entropy loss function.
    The formalization is as follows:
    \begin{equation}
        \mathcal L=\alpha \mathcal L_w(y_w,\hat{y}_w)+(1-\alpha)\mathcal L_d(y_d,\hat{y}_d)
    \end{equation}
where $\alpha$ is a hyperparameter, $\hat{y}_*$ is the predicted distribution of word or dependency,
and the loss terms $\mathcal L_w(y_w,\hat{y}_w)$ and $\mathcal L_d(y_d,\hat{y}_d)$ are given by (\ref{eq:lw}) and (\ref{eq:ld}), respectively:
    \begin{equation}
    \label{eq:lw}
        \mathcal L_w(y_w,\hat{y}_w) = -\sum_{i=1}^{|V|}y_w^{(i)}\log(\hat{y}_w^{(i)})
    \end{equation}
    \begin{equation}
    \label{eq:ld}
        \mathcal L_d(y_d,\hat{y}_d) = -\sum_{i=1}^{|D|}y_d^{(i)}\log(\hat{y}_d^{(i)})
    \end{equation}
    where $|V|$ and $|D|$ are the sizes of the word lookup table and the dependency lookup table respectively.

    \section{Experiments}

    \subsection{Settings}
    We conducted experiments on two datasets.
    First, we trained on a dataset of 1.25 million sentence pairs from the LDC corpora\footnote{LDC2002E18, LDC2003E07, LDC2003E14, Hansards portion of LDC2004T07, LDC2004T08, and LDC2005T06.} for Chinese-to-English sentence pairs.
    Then, we used NIST MT02 as the development set and NIST MT03, 04, 05, and 06 as the test sets.
    We also conducted experiments using 4.43 million sentence pairs from the WMT'14 for the English-to-German sentence pairs, with newstest2012 as the development set and newstest2013, newstest2014, and newstest2015 as the test sets.
    All languages use SpaCy \cite{spacy} for tokenization and syntactic dependency parsing.
    Finally, we always use a single sentence as a reference for evaluation using case-insensitive 4-gram BLEU score \cite{papineni-etal-2002-bleu}.

    For the hyperparameters of the model, all hidden states are set to 512 dimensions.
    The word embeddings and dependency embeddings dimensions for both the source and target languages are set to 512.
    Training continues until there is no improvement in the BLEU score on the development set for 5 consecutive epochs.

    \subsection{Baseline Systems}
    To evaluate our proposed method, we compared it with relevant NMT methods:

    \textbf{Chen et al. (2017a)} \cite{chen-etal-2017-improved}:
    Propose a tree-coverage model that makes attention depend on the syntax of the source language.

    \textbf{Chen et al. (2017b)} \cite{chen-etal-2017-neural}:
    By incorporating source dependency information to enhance source representations.

    \textbf{BahdanauNMT} \cite{bahdanau2014neural}:
    The standard NMT model uses an attention mechanism (global attention).

    \textbf{LuongNMT} \cite{luong-etal-2015-effective}:
    Following the work of \cite{bahdanau2014neural}, local attention is employed to improve the performance of NMT translation.

    \subsection{Evaluating SynCoNMT}

    Table \ref{tab:evaluating_model_ch_en} and Table \ref{tab:evaluating_model_en_de} respectively show the translation results of SynCoNMT proposed in Section \ref{sec:proposed_approach} on Chinese-to-English and English-to-German translation tasks.
    We implemented SynCoNMT-B and SynCoNMT-L based on the baseline models BahdanauNMT \cite{bahdanau2014neural} and LuongNMT \cite{luong-etal-2015-effective}, respectively, to demonstrate the effectiveness of the proposed approach.
    Although the syntax-based method proposed by Chen et al. (2017a) \cite{chen-etal-2017-improved} does not always outperform attention-based methods, it indicates that syntactic information is valuable for NMT.

    In the Chinese-to-English translation task, SynCoNMT improved the performance of BahdanauNMT and LuongNMT by an average of 1.28 and 0.80 BLEU points, respectively.
    This indicates that SynCoNMT can effectively enhance the translation performance of NMT.
    Similarly, in the English-to-German translation task, SynCoNMT improved the performance of BahdanauNMT and LuongNMT by an average of 1.61 and 0.75 BLEU points, respectively.
    This suggests that SynCoNMT is a robust method.

    Specifically, compared to the syntax-based method proposed by Chen et al. (2017a) \cite{chen-etal-2017-improved}, SynCoNMT achieved higher BLEU scores, indicating that incorporating syntactic information into CVNNs is valuable.

    It is noteworthy that despite LuongNMT being more effective than BahdanauNMT in the baseline, SynCoNMT performed better on BahdanauNMT than on LuongNMT.
    This suggests that syntactic embedding is more effective when considering the overall sentence information, as the syntactic tree is essentially a tree structure with a holistic nature.

    \begin{table}
        \centering
        \caption{The results of SynCoNMT on Chinese-to-English translation tasks.}
        \label{tab:evaluating_model_ch_en}
        \begin{tabular}{p{3cm}*{5}{>{\centering\arraybackslash}p{1.5cm}}}
            \toprule
            \textbf{Models}    & \textbf{MT03}  & \textbf{MT04}  & \textbf{MT05}  & \textbf{MT06}  & \textbf{AVG.}  \\
            \midrule
            Chen et al. (2017a) & 35.64          & 36.63          & 34.35          & 30.57          & 34.30          \\
            Chen et al. (2017b) & 35.91          & 38.73          & 34.18          & 33.76          & 35.64          \\
            BahdanauNMT        & 35.24          & 37.49          & 34.60          & 32.48          & 34.95          \\
            LuongNMT           & 35.57          & 37.85          & 34.93          & 32.74          & 35.27          \\
            \midrule
            SynCoNMT-B         & \textbf{36.27} & \textbf{39.02} & \textbf{35.71} & \textbf{33.91} & \textbf{36.23} \\
            SynCoNMT-L         & 36.09          & 38.97          & 35.65          & 33.57          & 36.07          \\
            \bottomrule
        \end{tabular}
    \end{table}

    \begin{table}
        \centering
        \caption{The results of SynCoNMT on English-to-German translation tasks.}
        \label{tab:evaluating_model_en_de}
        \begin{tabular}{p{3cm}*{3}{>{\centering\arraybackslash}p{2.3cm}} >{\centering\arraybackslash}p{1.8cm}}
            \toprule
            \textbf{Models}    & \textbf{newstest2013} & \textbf{newstest2014} & \textbf{newstest2015} & \textbf{AVG.} \\
            \midrule
            Chen et al. (2017a) & 20.78                 & 19.43                 & 20.37                 & 20.19          \\
            Chen et al. (2017b) & 20.91                 & 19.35                 & 20.57                 & 20.28          \\
            BahdanauNMT        & 20.23                 & 18.67                 & 19.78                 & 19.56          \\
            LuongNMT           & 20.74                 & 19.00                 & 20.15                 & 19.96          \\
            \midrule
            SynCoNMT-B         & \textbf{21.89}        & \textbf{20.72}        & \textbf{20.91}        & \textbf{21.17} \\
            SynCoNMT-L         & 21.43                 & 20.54                 & 20.17                 & 20.71          \\
            \bottomrule
        \end{tabular}
    \end{table}

    \subsection{Effect of Translating Long Sentences}

    \begin{figure}
        \includegraphics[width=\textwidth]{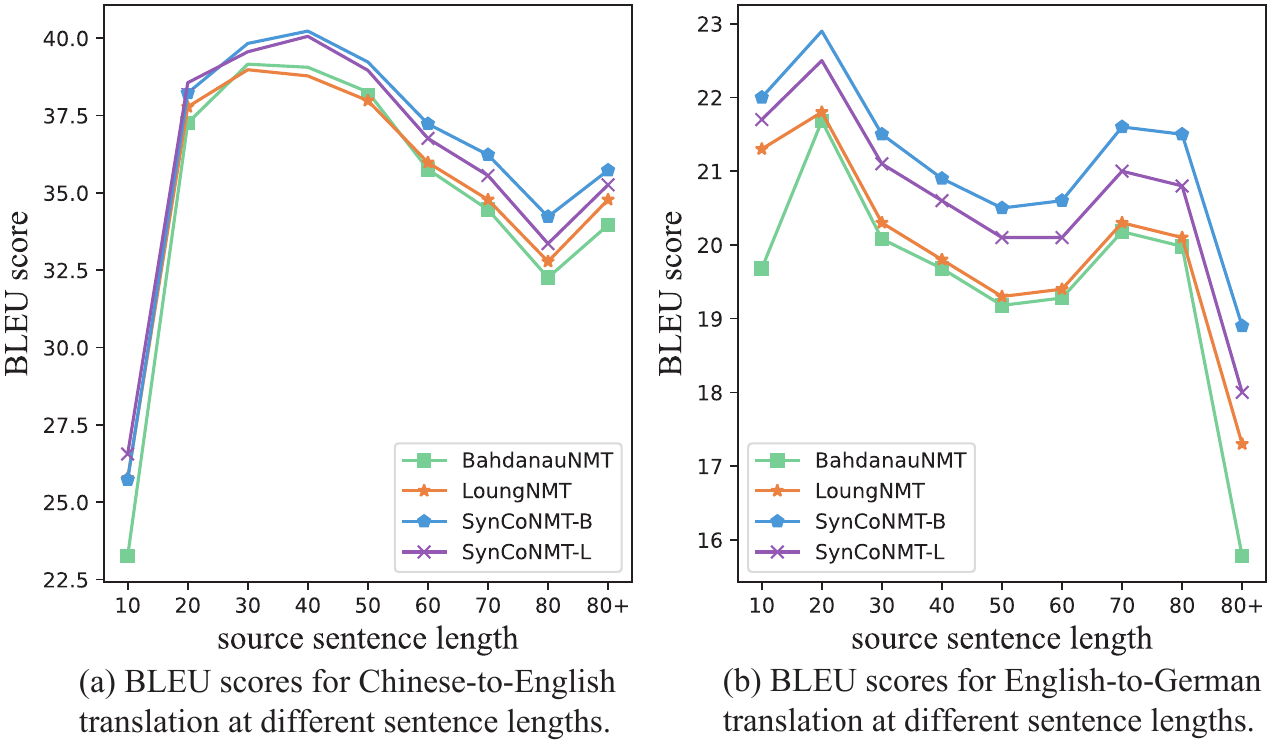}
        \caption{The performance of SynCoNMT on different sentence lengths.}
        \label{fig:sentence}
    \end{figure}

    We grouped similar-length sentences in the test sets of both tasks to evaluate BLEU performance.
    For instance, the sentence length ``30'' denotes sentences with source lengths between 20 to 30.
    Then, we calculated the BLEU score for each group, and the results are shown in Figure \ref{fig:sentence}(a) and Figure \ref{fig:sentence}(b).
    SynCoNMT consistently produced higher BLEU scores than the baseline BahdanauNMT and LuongNMT on sentences of different lengths in both translation tasks.
    Particularly, in terms of performance on longer sentences, SynCoNMT-B achieved significantly higher BLEU scores than other methods.
    This is because our method focuses more on syntactic matching between the source and target sentences, rather than just considering the syntactic structure of one side (either the source or target).
    Moreover, the use of complex embeddings allows for joint word and syntax constraints on the predicted output, improving the performance of NMT.

    \subsection{Complex-valued Attention}
    \begin{figure}
        \includegraphics[width=\textwidth]{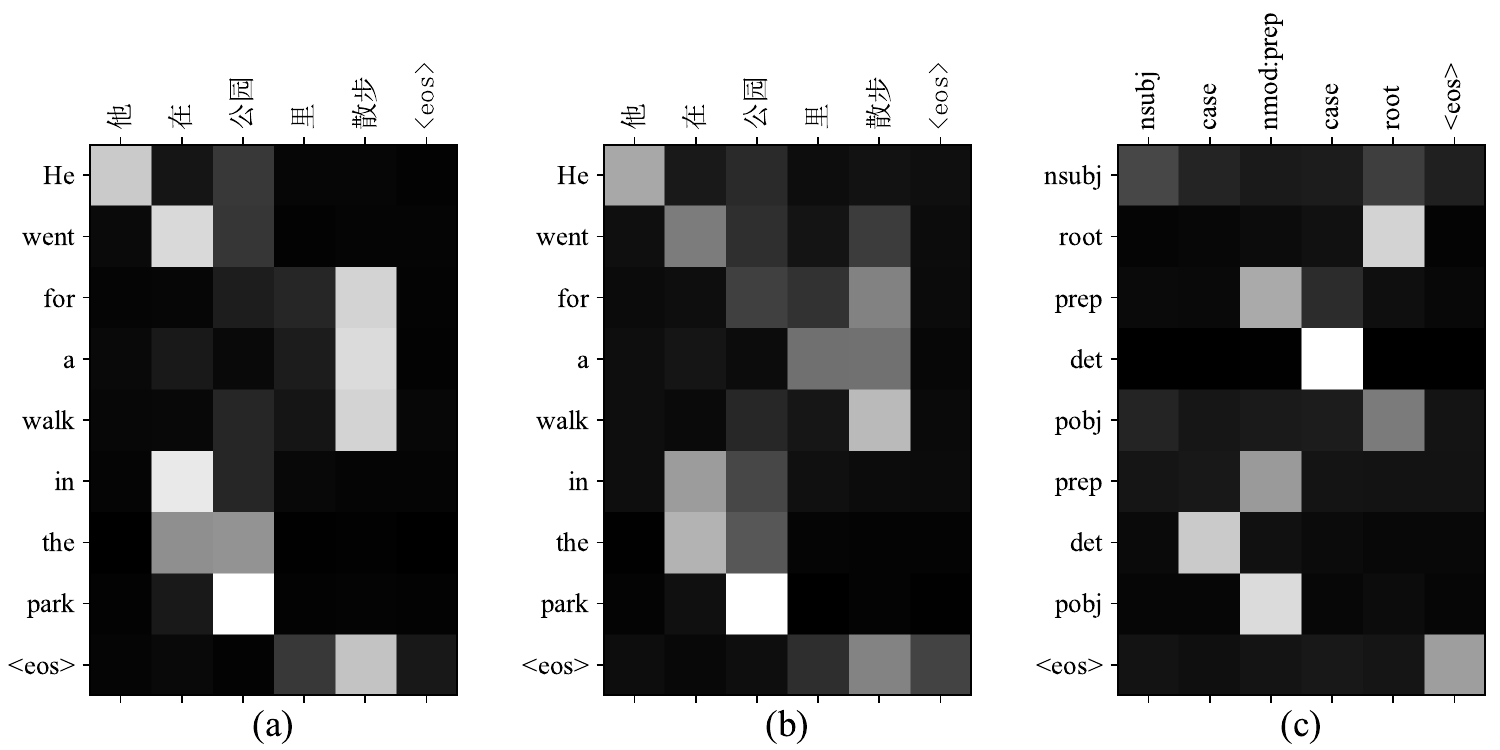}
        \caption{The x-axis and y-axis of each figure respectively represent the words or dependencies in the target language (English) and the source language (Chinese). (a) shows the results of traditional attention. (b) shows the real part of attention scores in SynCoNMT. (c) shows the imaginary part of attention scores in SynCoNMT.}
        \label{fig:attention}
    \end{figure}

    Figure \ref{fig:attention} illustrates an example where the complex attention scores differ from the traditional attention scores.
    At the sentence level, the real part of the complex-valued attention scores tends to focus more on the words that have syntactic dependencies with it.
    At the dependency level, the imaginary part of the complex-valued attention scores focus on the syntactic dependency relationships.
    By jointly weighting the complex embeddings, the model predicts both the target sentence and its corresponding syntax through the decoder. This further demonstrates the robustness of SynCoNMT.

    \section{Related Work}
    In this section, we will introduce the relevant works on Syntax-based NMT and CVNNs.

    Syntactic information has been shown to be effective in SMT \cite{li-etal-2013-modeling,williams-koehn-2014-syntax}.
    Therefore, a lot of work has been done on how to incorporate syntactic information into NMT.
    Using LSTM or GRU to model syntactic trees is a common practice \cite{miwa-bansal-2016-end,kokkinos-potamianos-2017-structural}.
    Hashimoto et al. \cite{hashimoto-tsuruoka-2017-neural} proposed combining head information with sequential words as input and modeling syntactic information using latent dependency graphs.
    Chen et al. \cite{chen-etal-2017-neural} used convolutional neural networks to represent dependency trees.
    Wu et al. \cite{ijcai2017p584} integrated syntactic information through multiple Bi-RNNs.
    Syntax-based NMT research is very active, but previous methods have focused on modeling syntactic information in the source language while ignoring syntactic information in the target language.

    CVNNs have received increasing attention in recent years because they can capture the interrelationships between real and imaginary parts within a single network.
    One of the early works on CVNNs is the complex backpropagation algorithm proposed by Hirose \cite{hirose1992continuous}.
    This algorithm extends the backpropagation algorithm to the complex domain, enabling the training of CVNNs.
    Inspired by Hirose's work, Zhang et al. \cite{zhang2017complex} proposed complex-valued convolutional neural networks.
    Popa et al. \cite{popa2018complex} proposed complex deep belief networks.
    Virtue et al. \cite{virtue2017better} proposed an approach for fingerprint recognition using CVNNs, which demonstrated better performance than its real-valued counterpart.
    Zhu et al. \cite{zhu2018quaternion} proposed a quaternion-based convolutional neural network and demonstrated its superiority in image classification and object detection tasks.

    However, to the best of our knowledge, CVNNs have not yet been applied to the Seq2Seq architecture of NMT.
    Therefore, inspired by the exciting work in CVNNs, we investigated the application of CVNNs in NMT.

    \section{Conclusion}
    In this paper, we present SynCoNMT, which employs complex-valued modeling of syntactic information to improve the performance of NMT.
    Our approach not only outperforms its real-valued counterpart but also outperforms other syntax-based NMT approaches.
    In addition, unlike approaches that only consider syntax on the source side, SynCoNMT considers syntactic mappings on both the source and target sides.
    Experiments and analysis demonstrate the effectiveness of our approach.
    In the future, we will explore how to apply our approach to other tasks.

%
%
%
%
\bibliographystyle{splncs04}
\bibliography{bibliography}
%
%
%
%
\end{document}